\documentclass[journal, 10pt]{article}

\usepackage{cite}

\usepackage{algorithmic}

\usepackage{url}

\usepackage{amssymb}
\usepackage{amsmath}
\usepackage{amsthm}
\usepackage[nooneline]{subfigure}
\usepackage{graphicx}
\usepackage{varwidth}
\usepackage{float}
\usepackage{fullpage} 

\newtheorem{thm}{Theorem}[section]

\begin{document}
\title{Variational Depth from Focus Reconstruction}
\author{Michael~Moeller, Martin Benning, Carola Sch\"onlieb, Daniel Cremers
\thanks{M.M. (corresponding author, email michael.moeller@in.tum.de) is with the Department of Mathematics, Technische Universit\"at M\"unchen, Boltzmannstrasse 3, 85748 Garching. M.B. is with the Institute of Mathematics and Image Computing (MIC), Universit\"at zu L\"ubeck, Maria-Goeppert-Str. 3, 23562 L\"ubeck, Germany. C.S. is with the Department of Applied Mathematics and Theoretical Physics, Centre for Mathematical Sciences, Wilberforce Road, Cambridge, CB3 0WA, United Kingdom. D.C. are with the Department of Computer Science, Technische Universit\"at M\"unchen, Boltzmannstrasse 3, 85748 Garching. }}

\maketitle


\begin{abstract}
This paper deals with the problem of reconstructing a depth map from a sequence of differently focused images, also known as \textit{depth from focus} or \textit{shape from focus}. We propose to state the depth from focus problem as a variational problem including a smooth but nonconvex data fidelity term, and a convex nonsmooth regularization, which makes the method robust to noise and leads to more realistic depth maps. Additionally, we propose to solve the nonconvex minimization problem with a linearized alternating directions method of multipliers (ADMM), allowing to minimize the energy very efficiently. A numerical comparison to classical methods on simulated as well as on real data is presented.
\end{abstract}


\section{Introduction}
\begin{figure}[h]
\begin{center}
\subfigure[{Visualization of the data as a cube.}]{\includegraphics[width=0.48\textwidth]{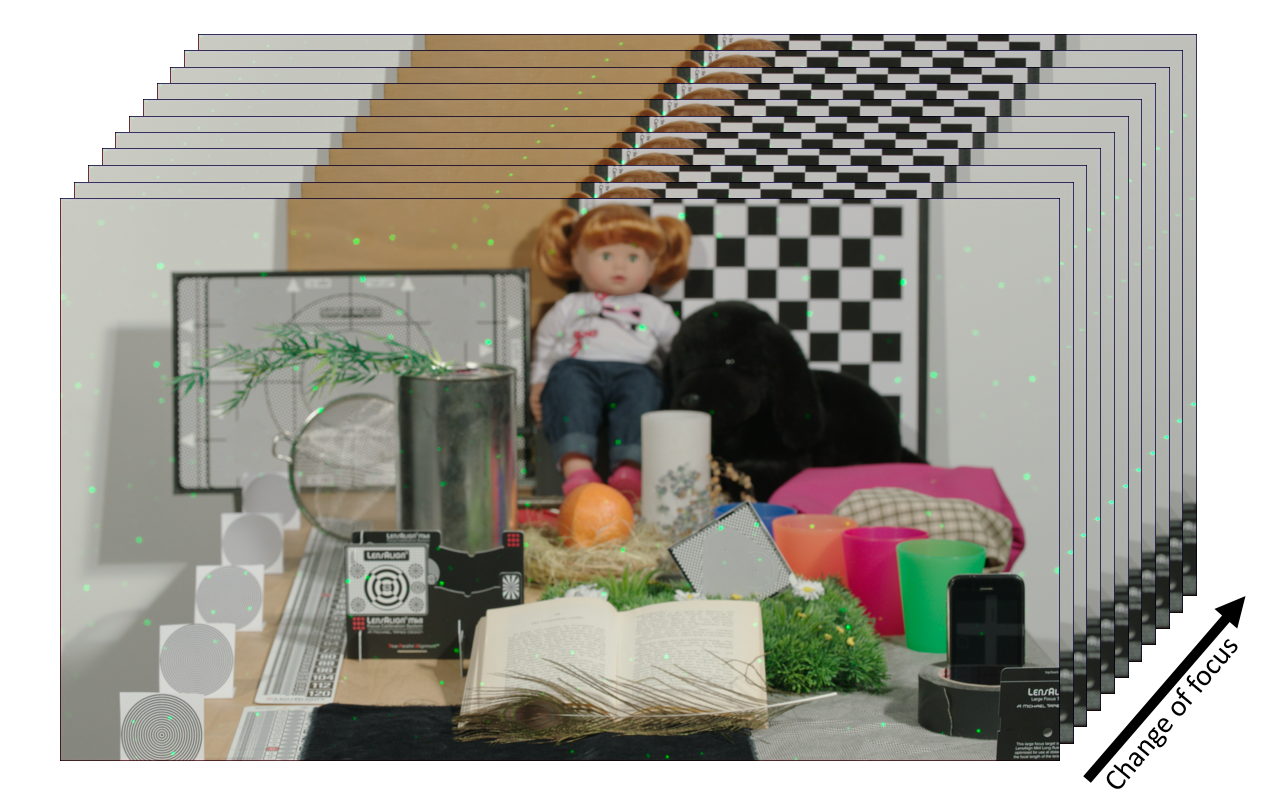}}
\subfigure[{Depth map reconstruction by finding the maximal contrast.}]{\includegraphics[width=0.48\textwidth, natwidth=560,natheight=420]{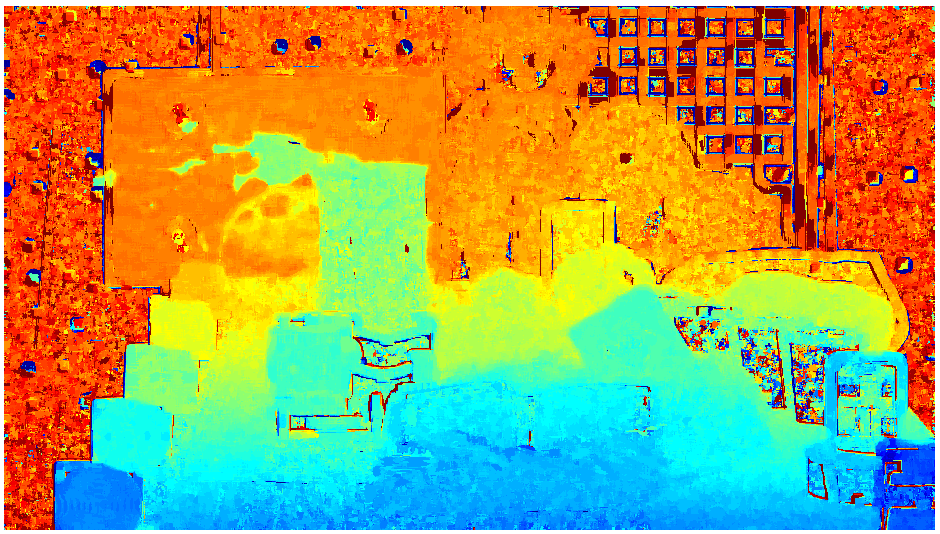}} 
\caption{Example of a simple DFF reconstruction: The data set of images can be visualized as a data cube (a), where the z-direction corresponds to a change of focus from the front to the back. In order to determine the focal setting at which the contrast is maximal, one picks a contrast measure, applies a windowed filtering to the contrast coefficients and selects the focal setting for which the coefficients are maximal. By knowing the distance at which a region appears maximally sharp in a focal setting, one reconstructs the depth. Figure (b) shows an example of the depth map, where red values indicate being far away from the camera and blue values correspond to pixels close to the camera. The result was obtained by using the modified Laplacian contrast measure with $9 \times 9$ windowed filtering as used in the comparison in \cite{Pertuz13Comparison}.}
\label{fig:commonApproach}
\end{center}
\end{figure}

The basic idea for \textit{depth from focus} (DFF) approaches is to assume that the distance of an object to the camera at a certain pixel corresponds to the focal setting at which the pixel is maximally sharp. Thus, for a given data set of differently focused images, one typically first finds a suitable contrast measure at each pixel. Subsequently, the depth at each pixel is determined by finding the focal distance at which the contrast measure is maximal. Figure \ref{fig:commonApproach} illustrates this approach. DFF differs from \textit{depth from defocus} (cf. \cite{paolo, Favaro08, Lin13}) in the sense that many images are given and depth clues are obtained from the sharpness at each pixel. Depth from defocus on the other hand tries to estimate the variance of a spatially varying blur based on a physical model and uses only very few images. Generally, the measurements of differently focused images do not necessarily determine the depth of a scene uniquely, such that the estimation of a depth map is an ill-posed problem. The ambiguity of the depth map is particularly strong in textureless areas.

The literature dealing with the shape from focus problem has very different contributions. A lot of work has targeted the development of different measures for sharpness and focus (cf. \cite{Pertuz13Comparison} for an overview). Other works deal with different methods of filtering the contrast coefficients before determining the maximum, i.e. the depth. The ideas range from windowed averaging (e.g. \cite{Thelen09}), over differently shaped kernels, to nonlinear filtering as proposed in \cite{Mahmood12}. In \cite{Muhammad09, Muhammad11}, the authors proposed to detect regions with a high variance in the depth map and suggested to smooth the depth in these parts by interpolation. Pertuz, Puig, and Garcia analyzed the behavior of the contrast curve in order to identify and remove low-reliability regions of the depth map in \cite{Pertuz13}. Ideas for using different focus measures and fusing the resulting depth estimates in a variational approach using the total variation (TV) regularization have been proposed in \cite{Mahmood13}. However, very little work has been done on finding a noisefree depth map in a single variational framework.

  
 Formulating the shape from focus problem in a variational framework has the advantage that one clearly defines a cost function and tries to find a solution which is optimal with respect to the costs. More importantly, regularity can be imposed on the depth estimate itself, e.g. by favoring piecewise constant solutions. Additionally, our framework is robust to noise and controls the ill-posedness of the shape from focus problem. 
 

While variational methods belong to the most successful class of methods for general image reconstruction tasks and have successfully been used in several depth from defocus approaches (c.f. \cite{Favaro10, Lin13, Liu10, Namboodiri08}), very little work has been done on using them for the DFF problem. The only work the authors are aware of is the method proposed in \cite{Gaganov09}, where the framework of Markov random fields was used to derive an energy minimization method consisting of two truncated quadratic functionals. However, using two nonconvex functionals in a setting where the dependency of the depth on the contrast is already nonconvex, results in a great risk of only finding poor local minima. For instance, any initialization with pixels belonging to both truncated parts is a critical point of such an approach.

This paper has two contributions: First, we propose a variational framework for the shape from focus problem using the total variation (TV) regularization \cite{ROF}. Secondly, we discuss the problem of minimizing the resulting nonconvex energy efficiently. While schemes that can guarantee the convergence to critical points are computationally expensive, we propose to tackle the minimization problem by an alternating minimization method of multipliers (ADMM) with an additional linearization of the nonconvex part.

\section{A Variational Approach}
\label{sec:model}
\subsection{Proposed Energy}
We propose to define a functional $E: \mathbb{R}^{n \times m} \rightarrow \mathbb{R}$ that maps a depth map $d \in \mathbb{R}^{n \times m}$ to an energy, where a low energy corresponds to a good depth estimate.  The dimension of the depth map, $n \times m$, coincides with the dimension of each image. The energy consists of two terms, $E = D + \alpha R$. The \textit{data fidelity term} $D$ takes the dependence on the measured data into account and the \textit{regularization term}  $R$ imposes the desired smoothness. The parameter $\alpha$ determines the trade-off between regularity and fidelity. We find the final depth estimate as the argument that minimizes our energy:
\begin{align}
\label{eq:energyMin}
\hat{d} = \arg \min_d D(d) + \alpha R(d).
\end{align}
Typical approaches from literature find the depth estimate by maximizing some contrast measure and we refer the reader to \cite{Pertuz13Comparison} for an overview of different contrast measures and their performance. Since we want to reconstruct the depth map by an energy minimization problem (rather than maximization) it seems natural to choose the negative contrast at each pixel as the data fidelity term,
\begin{align}
\label{eq:dataTerm}
D(d) = -\sum_i \sum_j c_{i,j}(d_{i,j}),
\end{align}
where $c_{i,j}$ denotes the (precomputed) function that maps a depth at pixel $(i,j)$ to the corresponding contrast value. With this choice our method is a generalization of methods that maximize the contrast at each pixel separately, since they are recovered by choosing $\alpha = 0$.

The regularization term $R$ imposes some smoothness on the recovered depth map and should therefore depend on the prior knowledge we have about the expected depth maps. In this paper we use the discrete isotropic TV, $ R(d) = \|Kd \|_{2,1}$,
 where $K$ is the linear operator such that $Kd$ is a matrix with an approximation to the $x$-derivative in the first column, to the $y$-derivative in the second column, and $\|g\|_{2,1} := \sum_{i} \sqrt{\sum_j (g_{i,j})^2}$.

\subsection{Contrast Measure}
\begin{figure}[H]
\begin{center}
\subfigure[{Contrast curve at green point}]{\includegraphics[width=0.4\textwidth]{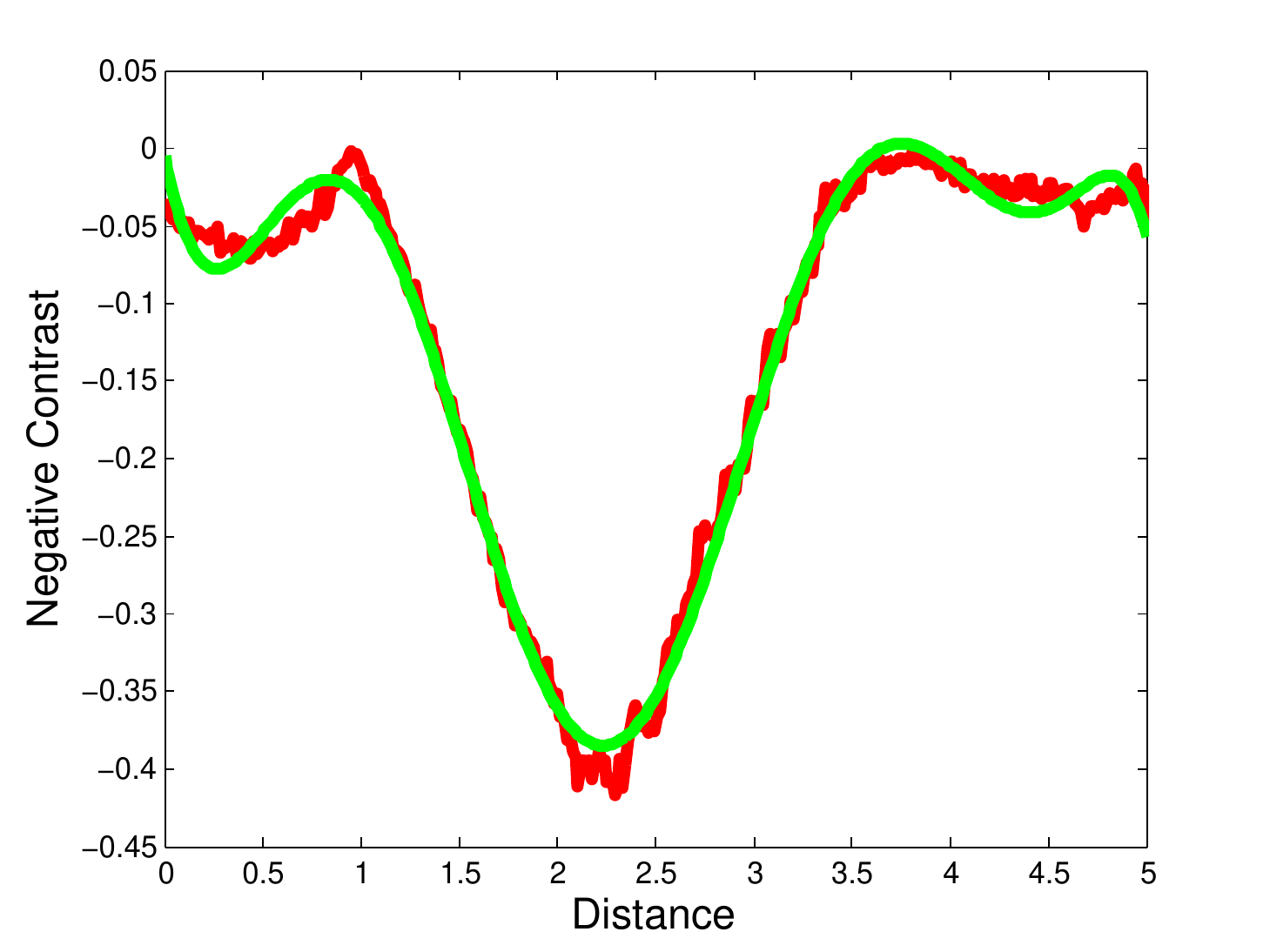}}
\subfigure[{Contrast curve at cyan point}]{\includegraphics[width=0.4\textwidth]{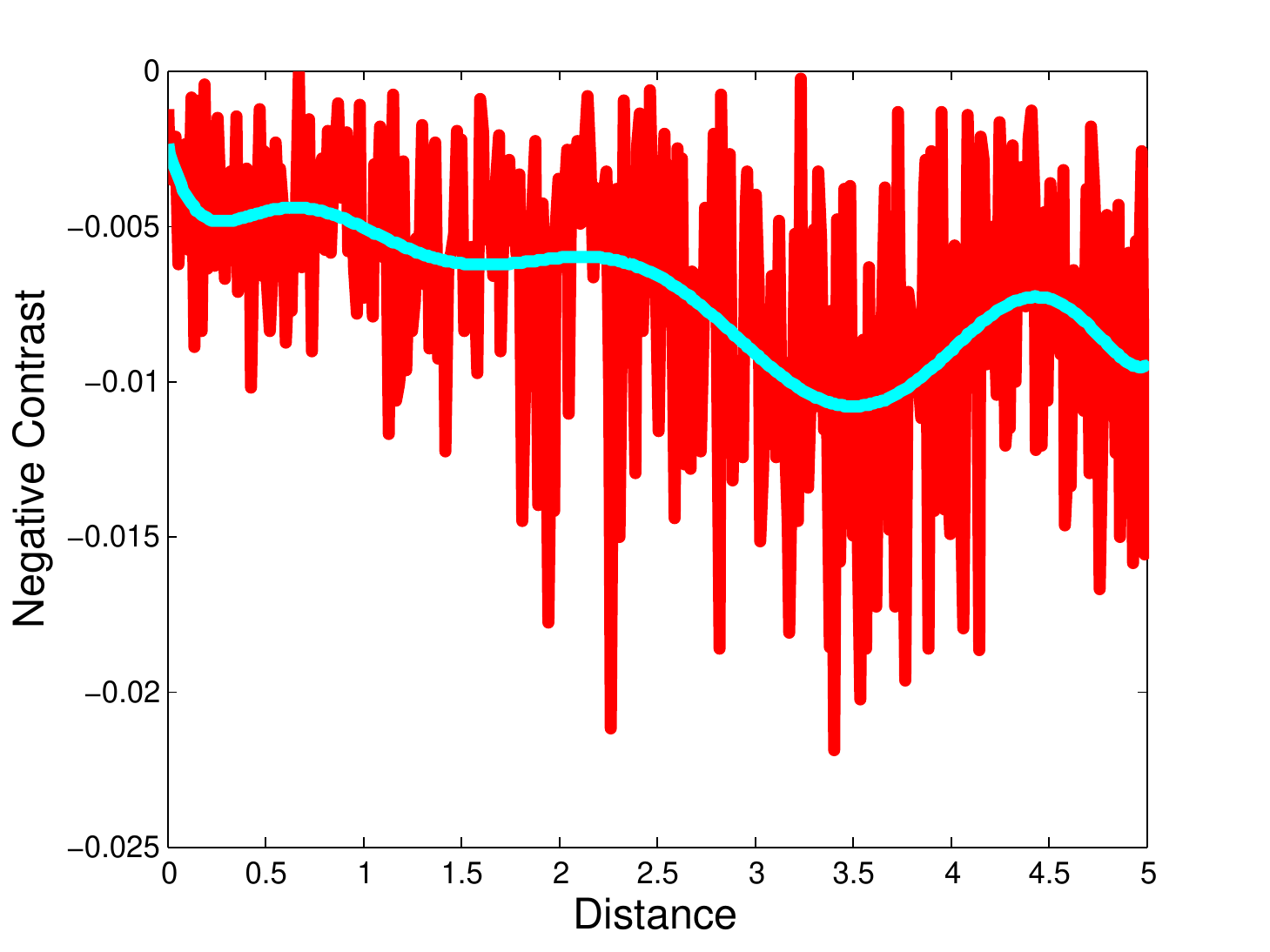}}\\
\subfigure[{Example image to illustrate the positions of the example points.}]{\includegraphics[width=0.8\textwidth, natwidth=560,natheight=420]{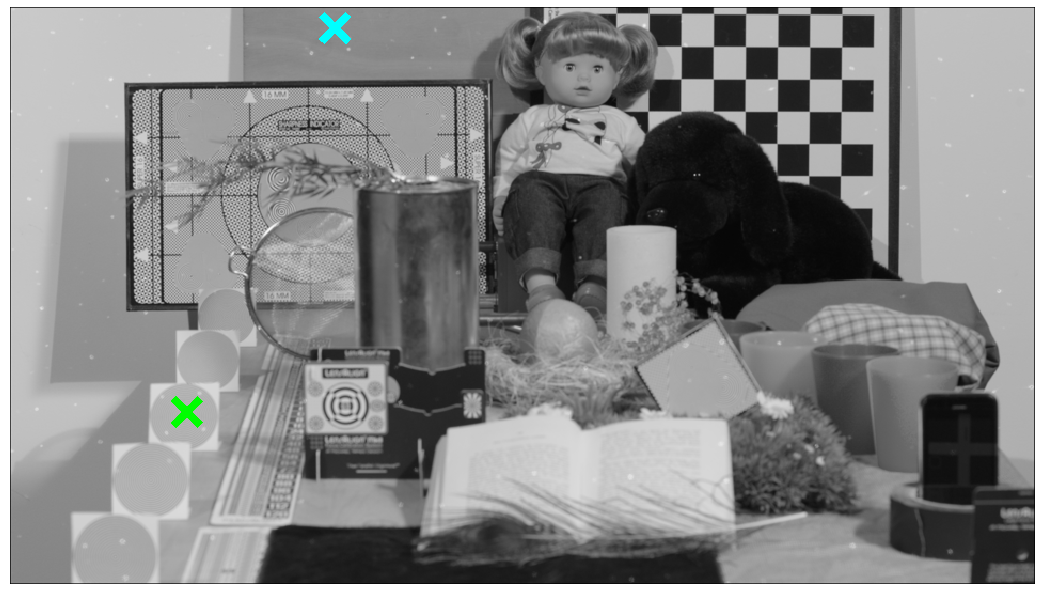}}
\caption{Examples for how the contrast curves look at different pixels. While pixels with a lot of surrounding structure (green point) result in a strong and clear contrast curve with an easy to identify minimum, textureless areas (cyan point) cause problems (as expected). }
\label{fig:contrastCurve}
\end{center}
\end{figure}

In this work, we choose the well-known modified Laplacian (MLAP) function \cite{Nayar94} as a measure of contrast. Note that under good imaging conditions the thorough study in \cite{Pertuz13Comparison} found that Laplacian based operators like the modified Laplacian consistently were among the contrast measures yielding the best performances. We determine
$$\text{MLAP}(i,j,k) = \sum_l |(\partial_{xx} I)(i,j,l,k)| + |(\partial_{yy} I)(i,j,l,k)| $$
with $(\partial_{xx} I)(i,j,l,k)$ denoting the approximate second derivative of the $l$-th color channel of the $k$-th image in the focus sequence in $x$-direction at pixel $(i,j)$ using a filter kernel $[1,\; -2,\;1]$. For each pixel $(i,j)$ we determine the continuous contrast function $c_{i,j}(x)$ depending on a depth $x$ by an eighth order polynomial approximation to the $\text{MLAP}(i,j,k)$ data. Polynomial approximations for determining the contrast curve have previously been proposed in \cite{Subbarao95} and offer a good and computationally inexpensive continuous representation of the contrast, which we need for the variational formulation \eqref{eq:dataTerm}. Different from \cite{Subbarao95} we use a higher order polynomial to approximate a noise free contrast curve using all data points, see Figure \ref{fig:contrastCurve} for an example. Obviously, areas with lots of texture are well suited for determining the depth. In areas with no texture (Figure \ref{fig:contrastCurve} (b)) the contrast measure consists of mostly noise. The magnitude of the contrast values is very low in this case, such that in a variational framework its influence will be small. 

We can see that the contrast values can be well approximated by smooth curves such as splines or - for the sake of simplicity in the minimization method - by a higher order polynomial. However, we can also see that even the smoothed contrast curves will not be convex. Thus, our data fidelity does not couple the pixels, and is smooth, but nonconvex. The regularization term on the other hand is convex, but couples the pixels and is nonsmooth, i.e. not differentiable in the classical sense. In the next section we will propose an efficient algorithm that exploits the structure of the problem to quickly determine depth maps with low energies. 

\section{Numerical Minimization}
\label{sec:minimization}
In the literature of nonconvex optimization, particularly the one related to image processing, there exist approaches for problems of the form smooth nonconvex plus nonsmooth convex that provably convergence to critical points under mild conditions. Such approaches include for instance forward-backward splittings (FBS), see \cite{attouch13, iPiano} and the references therein, and methods based on the difference of convex functions (DC), e.g. \cite{phamdinh}. The drawback of these methods is that they require to solve a problem like TV denoising at each iteration which makes these schemes computationally expensive. A simple FBS approach that provably converges to a critical point would be 
$$ d^{k+1} = \arg \min_d \frac{1}{2}\|d - d^k + \tau \nabla D(d^k) \|^2  + \tau \alpha R(d).$$
While one avoids the difficulty of dealing with the nonconvex term by linearizing it, it can be seen that one still has to minimize the sum of a quadratic fidelity and a nonsmooth regularization term, which can be costly. Therefore, we propose to apply the alternating directions method of multipliers (ADMM) (cf. \cite{boyd11}) in the same way as if the energy was convex. We introduce a new variable $g$ along with the constraint $g = K d$, such that we can rewrite the energy minimization problem \eqref{eq:energyMin} as
\begin{align*}
(\hat{d},\hat{g}) = \arg \min_{d,g} D(d) + \alpha \|g\|_{2,1} \ \ \text{such that } g = Kd.
\end{align*}
The constraint $g=Kd$ is enforced iteratively by using the augmented Lagrangian method. The minimization for $g$ and $d$ is done in an alternating fashion such that a straight forward application of the ADMM would yield
\begin{subequations}
\begin{align}
\label{eq:ADMM1}
d^{k+1} =& \arg \min_d \frac{\lambda}{2} \|Kd - g^k + b^k\|_2^2 + D(d), \\
\label{eq:ADMM2}
g^{k+1} =& \arg \min_g  \frac{\lambda}{2} \|g - Kd^{k+1} + b^k\|_2^2 + \alpha \|g\|_{2,1}, \\
\label{eq:ADMM3}
b^{k+1} =& b^k + (Kd^{k+1} - d^{k+1}).
\end{align}
\end{subequations}
Due to the nonconvexity of $D(d)$ subproblem \eqref{eq:ADMM1} is still difficult to solve, which is the reason why we additionally incorporate the FBS idea of linearizing the nonconvex term. The final computational scheme becomes
\begin{equation}
\label{eq:linearizedADMM}
\begin{aligned}
d^{k+1} =& \arg \min_d \frac{\lambda}{2} \|Kd - g^k + b^k\|_2^2 \\ 
&\qquad \ \ \  + \frac{1}{2}\|d - d^k + \tau \nabla D(d^k) \|^2, \\
g^{k+1} =& \arg \min_g  \frac{\lambda}{2} \|g - Kd^{k+1} + b^k\|_2^2 + \alpha \tau \|g\|_{2,1}, \\
b^{k+1} =& b^k + (Kd^{k+1} - d^{k+1}).
\end{aligned}
\end{equation}

Note that now each subproblem can be solved very efficiently. The update for $d$ involves the inversion of the operator $(\lambda K^T K + I)$ (a discretization of $I - \lambda \Delta$) which can be done using a discrete cosine transform. The update for $g$ has a closed form solution known as soft shrinkage of \mbox{$z = Kd^{k+1} - b^k$}, $$g_{i,j} = \frac{z_{i,j}}{\|z_{i,:}\|_2} \cdot \max(\|z_{i,:}\|_2 - \alpha \tau / \lambda, 0).$$

In this sense, the proposed algorithm can also be interpreted as solving each FBS subproblem with a single ADMM iteration (and keeping the primal and dual variables). We simply omit converging to the minimum of the convex energy but compute a new linearization of the data fidelity term at each iteration. From a different point of view, the above could also be interpreted as a generalization of Bregmanized operator splitting \cite{zhang}. In the context of the literature on inverse problems the linearization is related to the Levenberg-Marquardt method. 
Although the techniques contributing to the algorithm are all known, the authors are not aware of any literature using the method as stated above - even in the convex setting. Hence, let us briefly state a convergence result in the convex setting:
\begin{thm}
\label{thm:convexConvergence}
If $D$ and $R$ are convex and the symmetric Bregman distance with respect to $D$, $S_D(d,v) = \langle d - v, p_d - p_v\rangle $ with $p_d \in \partial D(d)$, $p_v \in \partial D(v)$, can be bounded by $\frac{1}{2\tau}\|d-v\|^2$, then the iterates $d^k$ produced by \eqref{eq:linearizedADMM} converge to a solution $\hat{d}$ of \eqref{eq:energyMin} with
\begin{align*}
\tau \left(S_D(p^{k}, \hat{p})+\alpha S_R(g^{k},K\hat{d}) \right) \leq C/k
\end{align*}
and $\|Kd^{k+1}-g^k\|^2 \leq C/k$ for some constant $C$.
\end{thm}
The proof follows the arguments of Cai et al. in \cite{SplitBregmanConvergence}, and is given in the appendix. 

While the idea of applying splitting methods to nonconvex problems as if they were convex is not new, a convergence theory is still an important open problem. Recently, progress has been made for particular types of nonconvexities, e.g. \cite{Valkonen14, M14}. One result we became aware of after the preparation of this manuscript is the preprint \cite{Li14} by Li and Pong. They investigate the behavior of ADMM algorithms for nonconvex problems and even include the linearization proposed in this manuscript. They prove the convergence of the algorithm to stationary points under the assumption that the nonconvex part is smooth with bounded Hessian and that the linear operator ($K$ in our notation) is surjective. Unfortunately, the latter is not the case for our problem. However, the proposed method numerically results in a stable optimization scheme that is more efficient than methods that solve a full convex minimization problem at each iteration. 

Another class of methods related to our approach are quadratic penalty methods, cf. \cite{nikolova10}, which we would obtain by keeping $b^k$ fixed to zero and steadily increasing $\lambda$. Variations exist where the alternating minimizations are repeated several times before increasing the penalty parameter $\tau$, see e.g. \cite{HyperLaplacian}. We refer the reader to \cite{attouch10} for a discussion on the convergence analysis for this class of methods. While we find a steadily increasing parameter $\lambda$ to improve the convergence speed, we observe that including the update in $b^k$, i.e. doing ADMM instead of a quadratic penalty, leads to a much faster algorithm. 

\section{Numerical Results}
\subsection{Experiments on Simulated Data}
The numerical experiments are divided into three parts. In the first part, we test the proposed framework on simulated data with known ground truth and compare it to classical depth from focus methods. We use the Matlab code provided at \url{http://www.sayonics.com/downloads.html} by the authors of \cite{Pertuz13} and \cite{Pertuz13Comparison} for the simulation of the defocused data as well as for computing depth maps. The simulation can take an arbitrary input image (typically some texture) and creates a sequence of images (15 in our examples) that are focus blurred as if the textured had a certain shape. A mixture of intensity dependent and intensity independent Gaussian white noise is added to the images to simulate realistic camera noise. We refer to \cite{Pertuz13Comparison} as well as to the implementation and demonstration at  \url{http://www.sayonics.com/downloads.html} for details. The classical depth from focus code allows to choose different contrast measures and different filter sizes, i.e. different mean filters applied to the contrast values before the actual depth is determined. The depth can have subpixel accuracy by using the default 3p Gaussian interpolation. The code additionally allows to apply a median filter to the final depth map to further average out errors. Since the objective of this paper is the introduction of variational methods to the depth from focus problem we limit our comparison to the modified Laplacian (MLAP) contrast measure for the variational as well as for the classical approach and just vary the filtering strategies. 

\begin{figure}[h]
\begin{center}
\includegraphics[width=1\textwidth]{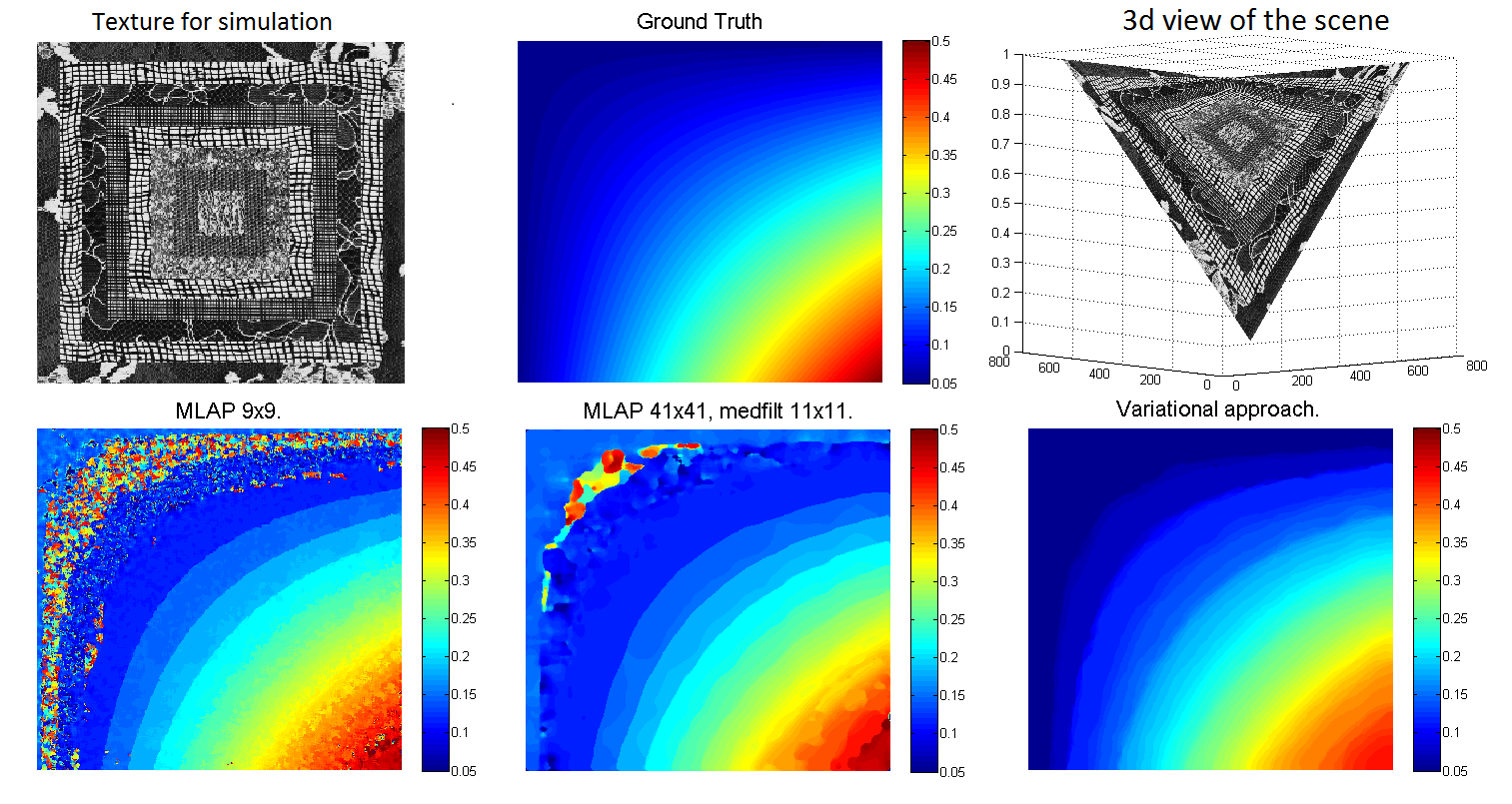}
\caption{Comparison of different depth from focus methods for simulated data. Top row from left to right: Texture image used for the simulation, true simulated depth map, 3d view of the simulated scene.  Bottom row from left to right: MLAP depth estimate with $9\times 9$ windowed filtering, MLAP depth estimate with $41\times 41$ windowed filtering and an additional $11\times 11$ median filter on the resulting depth image, our proposed variational scheme with regularization parameter $\alpha = 1/2$.  The mean squared error (MSE) is given for each of the depth map reconstructions.}
\label{fig:simulatedComparison2}
\end{center}
\end{figure}

We set up the variational model as proposed in Section \ref{sec:model}, and minimize it computationally as described in Section \ref{sec:minimization}. Figure \ref{fig:simulatedComparison2} shows a comparison of our framework with differently filtered depth from focus methods along with the mean squared error (MSE) to the ground truth for a stack of differently focused images using the texture in the upper left of Figure \ref{fig:simulatedComparison2} for the simulation.

%
%

As we can see the variational approach behaves superior to the classical ones. The image used to simulate the results shown in Figure \ref{fig:simulatedComparison2} is textured almost everywhere such that parts with high contrast can often be identified accurately. However, the simulated depth map changes quickly, which is the reason why there exists a region which does not appear perfectly sharp in any of the 15 simulated images. As we can see in Figure \ref{fig:simulatedComparison2} this region causes big problems and even a $41\times41$ contrast filtering followed by an $11\times 11$ median filtering cannot remove the erroneous parts completely. Moreover, strong filtering can only be applied without visible loss of details if the underlying depth map is smooth. The latter holds for the simulated depth maps used for the comparison, but is unrealistic for real data. The problem of unreliable regions has been addressed in \cite{Pertuz13} before, however, by making a binary decision for or against the reliability of each pixel. Our variational approach handles unreliable regions automatically, in the sense that they do not show large contrast maxima such that their influence on the total energy is small. This allows the method to change the depth at these pixels at low costs and, in this sense, leads to a fuzzy instead of a binary decision for the reliability of pixels. 

Additional to the simulation in Figure \ref{fig:simulatedComparison2}, we chose three more types of depth maps for our simulated data. The root mean square error values (RMSE) in Table \ref{tab:rmse} show that the variational approach showed superior performance for all simulated shapes.

\begin{table}[H]
\begin{center}
\begin{tabular}{ |p{2cm}| c| c| c| c|}
\hline
 Method / Shape & Cone & Plane & Cosine & Sphere  \\\hline
MLAP 1 & 2.51& 9.87& 4.07& 8.64 \\\hline
MLAP 2 & 1.41& 5.14& 1.83& 5.06 \\\hline
TV, $\alpha = 1$ & 1.45& \textbf{1.01}& 1.73& 0.96 \\\hline
TV, $\alpha = 1/4$&\textbf{ 1.23}& 1.15& \textbf{1.24}& \textbf{0.89} \\\hline
TV, $\alpha = 1/8$ & 1.27 & 1.26 & 1.43& 0.97 \\\hline 
\end{tabular}
\label{tab:rmse}
\caption{RMSE values different methods obtained on the simulated depth maps. MLAP 1 denotes a $9 \times 9$ windowed filtering of the MLAP contrast coefficients. MLAP 2 denotes a $41 \times 41$ windowed filtering of the MLAP contrast coefficients followed by a $11\times 11$ median filter on the depth map. TV denotes our variational approach using TV regularization for different parameters.}

\vspace{-0.7cm}
\end{center}
\end{table}
Smooth surfaces are not very realistic scenarios for depth maps in practice, because they do not contain jumps and discontinuities. Therefore, we compare the different methods on real data in the next section. 

\subsection{Experiments on Real Data}
We test the proposed scheme on two data sets recorded with different cameras. One was downloaded from Said Pertuz' website, \url{http://www.sayonics.com/downloads.html}, and consists of 33 color images with $640\times 480$ pixels resolution captured with a Sony SNC RZ50P camera, $f = 91$mm. 

\begin{figure}[H]
\begin{center}
\begin{tabular}{p{0.31\textwidth}p{0.31\textwidth}p{0.31\textwidth}}
\includegraphics[width=0.31\textwidth]{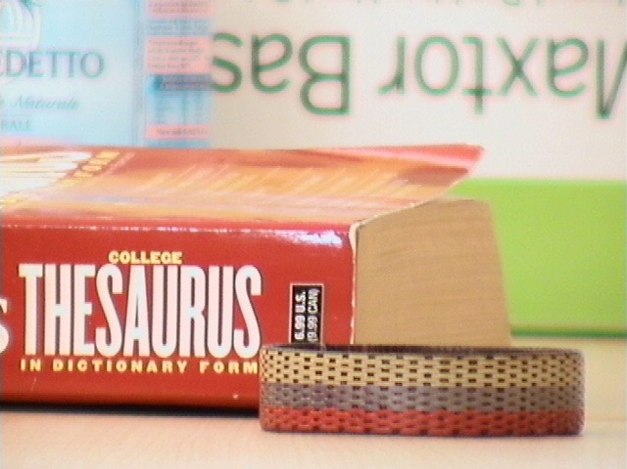}  & 
\includegraphics[width=0.31\textwidth]{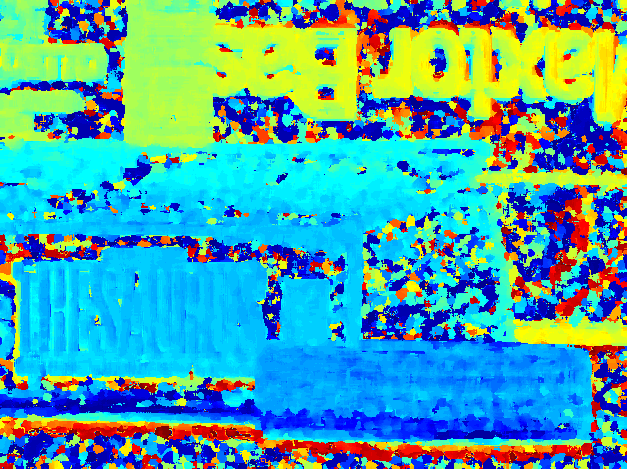}&
\includegraphics[width=0.31\textwidth]{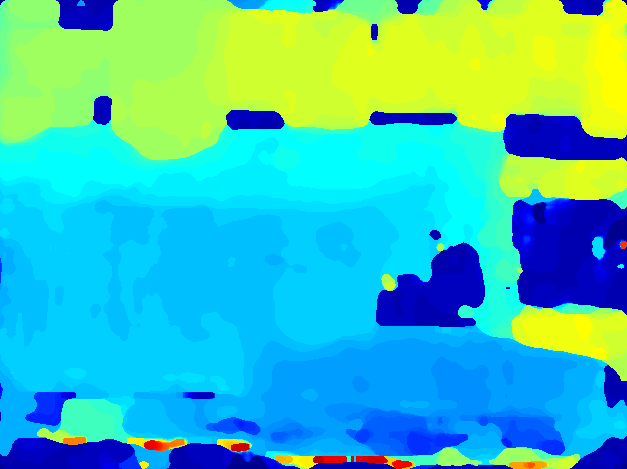} \\
(a) Example image & (b) $9 \times 9$ windowed MLAP maximization.  &  (c) $41 \times 41$ windowed MLAP maximization with $11\times 11$ median filter.\\
\includegraphics[width=0.31\textwidth]{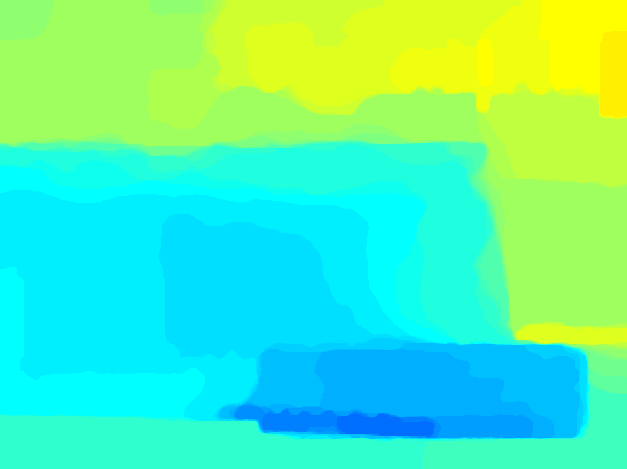}  & 
\includegraphics[width=0.31\textwidth]{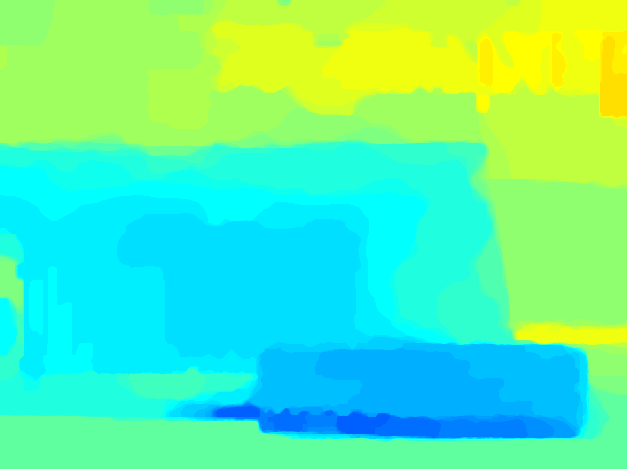}&
\includegraphics[width=0.31\textwidth]{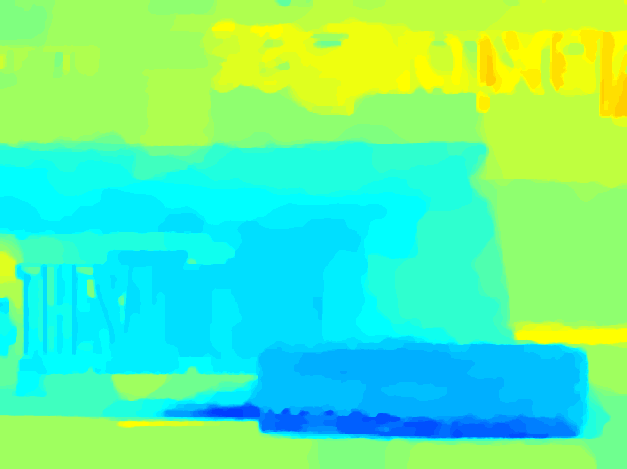} \\
(d) Variational approach, \mbox{$\alpha = 1/5$}.& (e) Variational approach, \mbox{$\alpha = 1/7$}. &  (f) Variational approach, \mbox{$\alpha = 1/10$}.
\end{tabular}
\caption{Comparison of depth from focus reconstruction on real data for differently filtered classical methods and our variational approach for different regularization parameters.}
\label{fig:variationalApproach}
\end{center}
\end{figure}

The results obtained by two differently filtered classical approaches as well as by using the variational method with three different regularization parameters is shown in Figure \ref{fig:variationalApproach}. As we can see, the variational approaches yield smoother and more realistic depth maps. 
 Even with very large filterings the MLAP depth estimate still contains areas that are erroneously determined to be in the image foreground. Depending on the choice of regularization parameter the results of our variational approach show different levels of details and different amounts of errors. However, even for the smallest regularization parameter no parts of the image are estimated to be entirely in the fore- or background. Generally, due to the rather low quality of the data and large textureless parts the estimation of the depth is very difficult on these images.

For our second test, we use high quality data taken with the ARRI Alexa camera\footnote{Provided by the Arnold and Richter Cine Technik, {www.arri.com}}. The latter data set consists of 373 images of a tabletop scene with an original resolution of $1080\times 1920$ pixels which we downscaled by a factor of two to reduce the amount of data. To record the image stack, the focus was continuously and evenly changed from the front to the back of the scene using a wireless compact unit (WCU-4) connected to a motor for focus adjustments. The images ran through the usual ARRI processing chain, which drastically influences the data characteristics as observed in \cite{Andriani13}. 

\begin{figure}[H]
\begin{center}
\begin{tabular}{p{0.47\textwidth}p{0.47\textwidth}}
\includegraphics[width=0.47\textwidth]{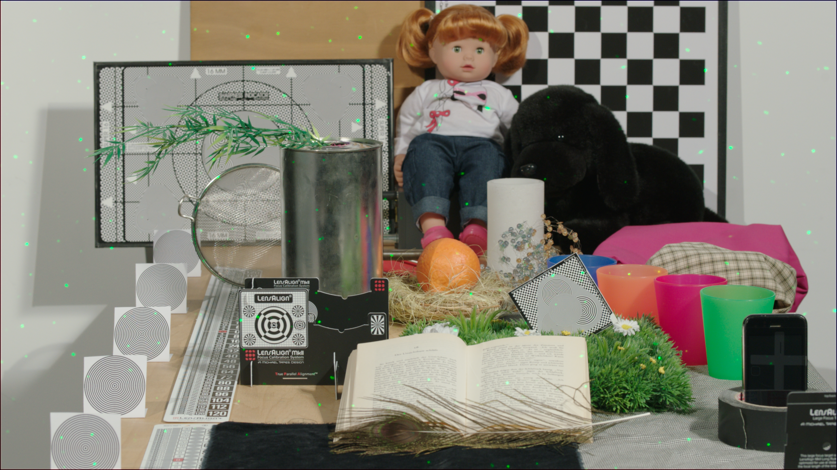}  & 
\includegraphics[width=0.47\textwidth]{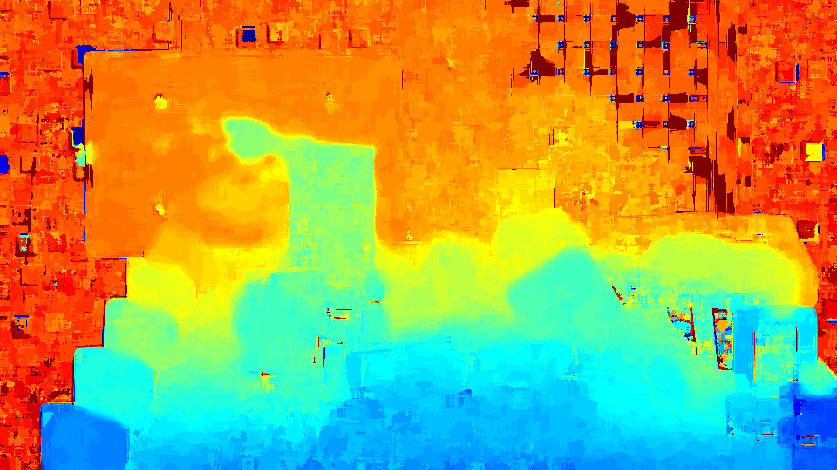} \\
(a) Example for one image of the focus sequence. & (b) Depth map obtained after filtering MLAP coefficients with a $21 \times 21$ window and applying a $5\times 5$ median filter to the resulting depth map.  \\
\includegraphics[width=0.47\textwidth]{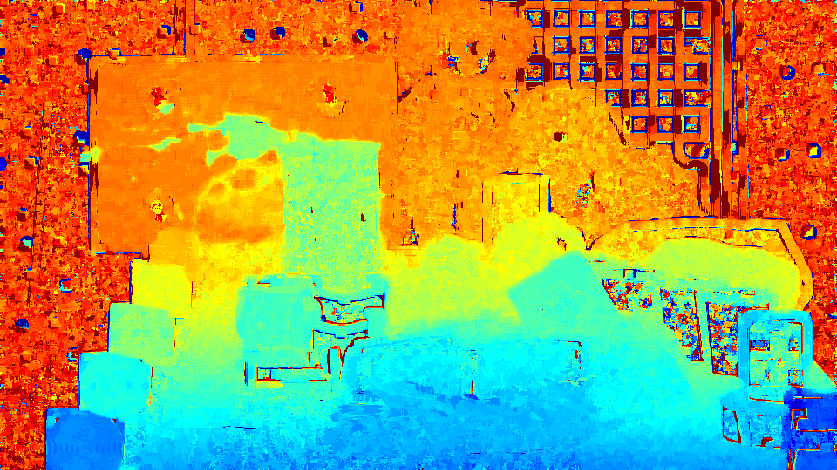} & 
\includegraphics[width=0.47\textwidth]{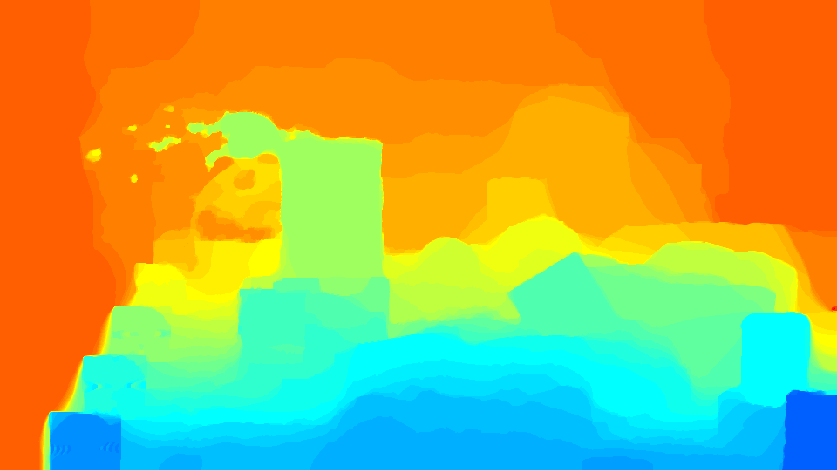} \\
(c) Depth map obtained after filtering MLAP coefficients with a $9 \times 9$ window.  & (d) Variational approach with $\alpha = 1/12$.\\
 \includegraphics[width=0.47\textwidth]{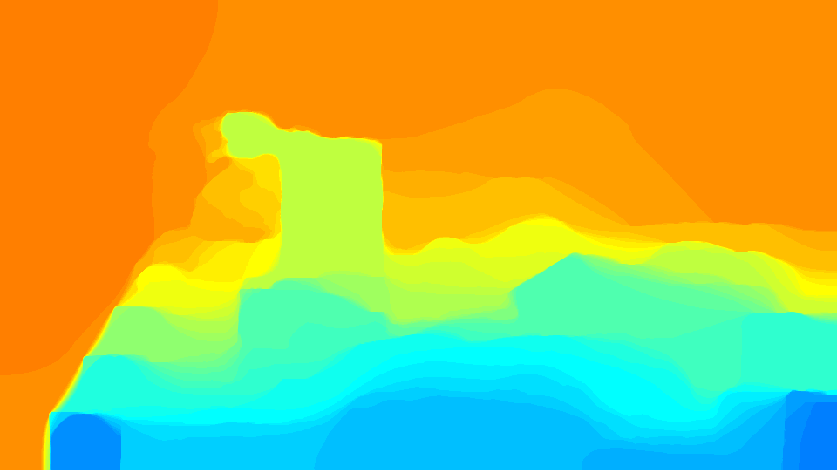}  & 
\includegraphics[width=0.47\textwidth]{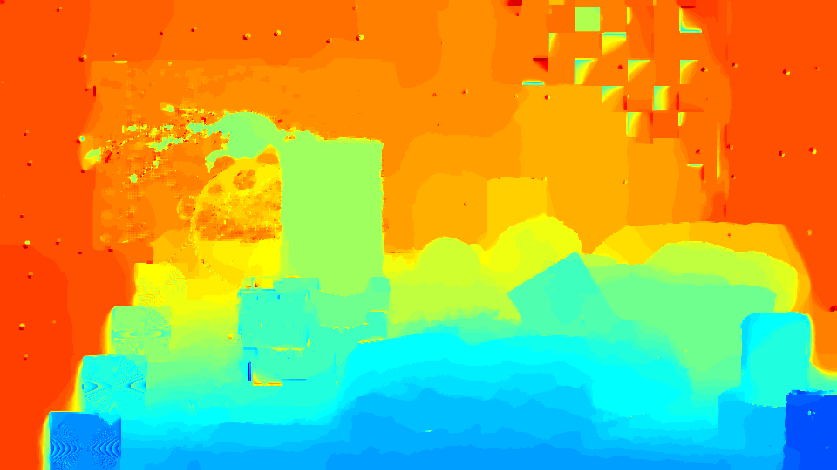} \\
(e) Variational approach with $\alpha = 1/4$.& (f) Variational approach with $\alpha = 1/28$.\\
  \includegraphics[width=0.47\textwidth]{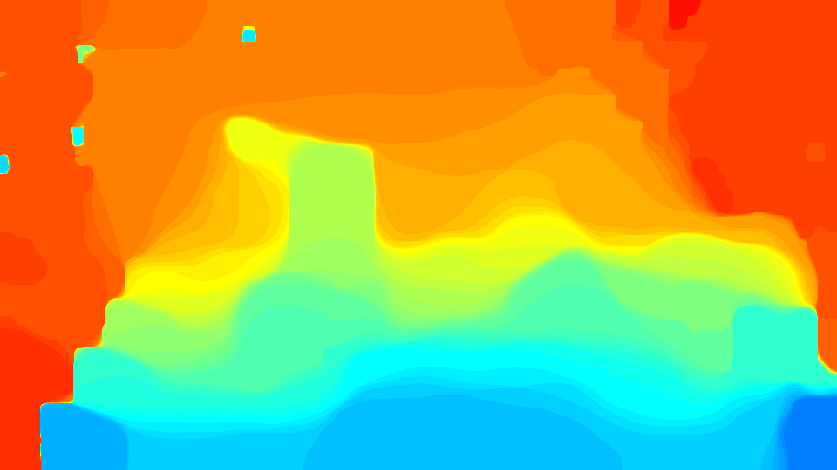}  & 
\includegraphics[width=0.47\textwidth]{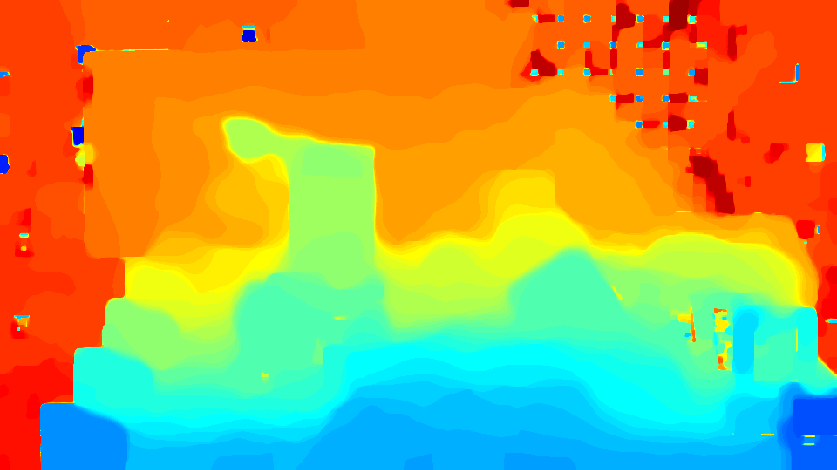} \\
(g) Applying TV denoising to image (b) with a regularization parameter of 26. & (h) Applying TV denoising to image (b) with a regularization parameter of 6.  \\
\end{tabular}
\caption{Comparison of depth from focus reconstruction on real data for differently filtered classical methods and our variational approach for different regularization parameters.}
\label{fig:variationalApproach2}
\end{center}
\end{figure}

As we can see in the results shown in Figure \ref{fig:variationalApproach2}, the depth estimation on the second, high quality data set works much better. Both MLAP methods succeed in finding the general structure of a reasonable 
\noindent depth map. However, obvious errors can be seen at some parts of the wall, the checkerboard and some transitions between objects on the table. The variational approaches succeed in finding a much more realistic solution by eliminating all noise in the depth image. Particularly impressive is the fact that they even restore the wall in the background, the checkerboard and the plastic cups in the foreground. All these things are not reconstructed correctly in the windowed approaches. As expected, the regularization parameter allows the user to determine the trade-off between keeping little details and suppressing noise.

In addition to the results of two classical and three variational approaches, Figure \ref{fig:variationalApproach2} shows the results obtained by first reconstructing the depth map and then applying TV denoising to the resulting depth map. While this procedure has the advantage that the TV problem is convex, it treads the estimation of the depth independent of the denoising. As we can see, our proposed joint approach for denoising a depth map with small TV works much better. Large parts that have erroneously been mapped to the foreground are very difficult to remove when applying TV denoising to the depth map and large regularization parameters are required. For the joint variational approach, mapping parts from the fore- to the background and vice versa is rather cheap as long as the contrast values do not show strong preferences for one over the other. Thus, the joint reconstruction is better suited for recovering realistic, noise-free depth maps. 

\subsection{Convergence of the Proposed Algorithm}
For the experiments shown in the previous subsection we used the minimization scheme described in Section \ref{sec:minimization} which we initialized with the $15 \times 15$ filtered MLAP depth estimate that have been additionally blurred with a $21 \times 21$ mean filter. The latter was done due to the observation that the minimization with false initializations that attains extreme values performs much worse than blurriness in a moderate range of depth values. As for the ADMM parameter we start with $\lambda = 1$ and increase the parameter at each iteration by $2\%$. Since we are using the scaled form of ADMM we consequently have to divide $b^k$ by $1.02$ at each iteration (see \cite{boyd11} for details).

\begin{figure}[ht]
\begin{center}
{\includegraphics[width=0.32\textwidth]{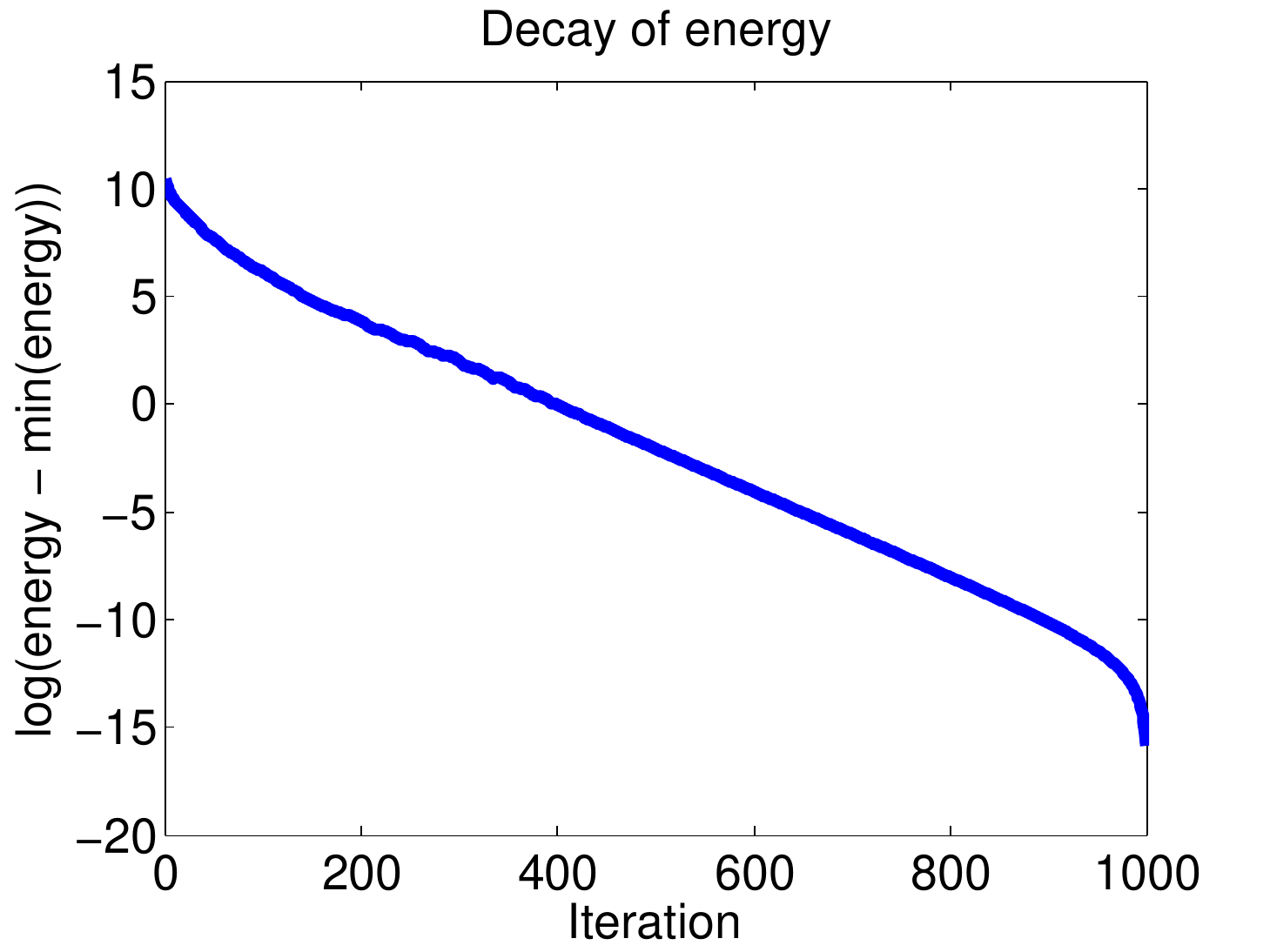}}
{\includegraphics[width=0.32\textwidth]{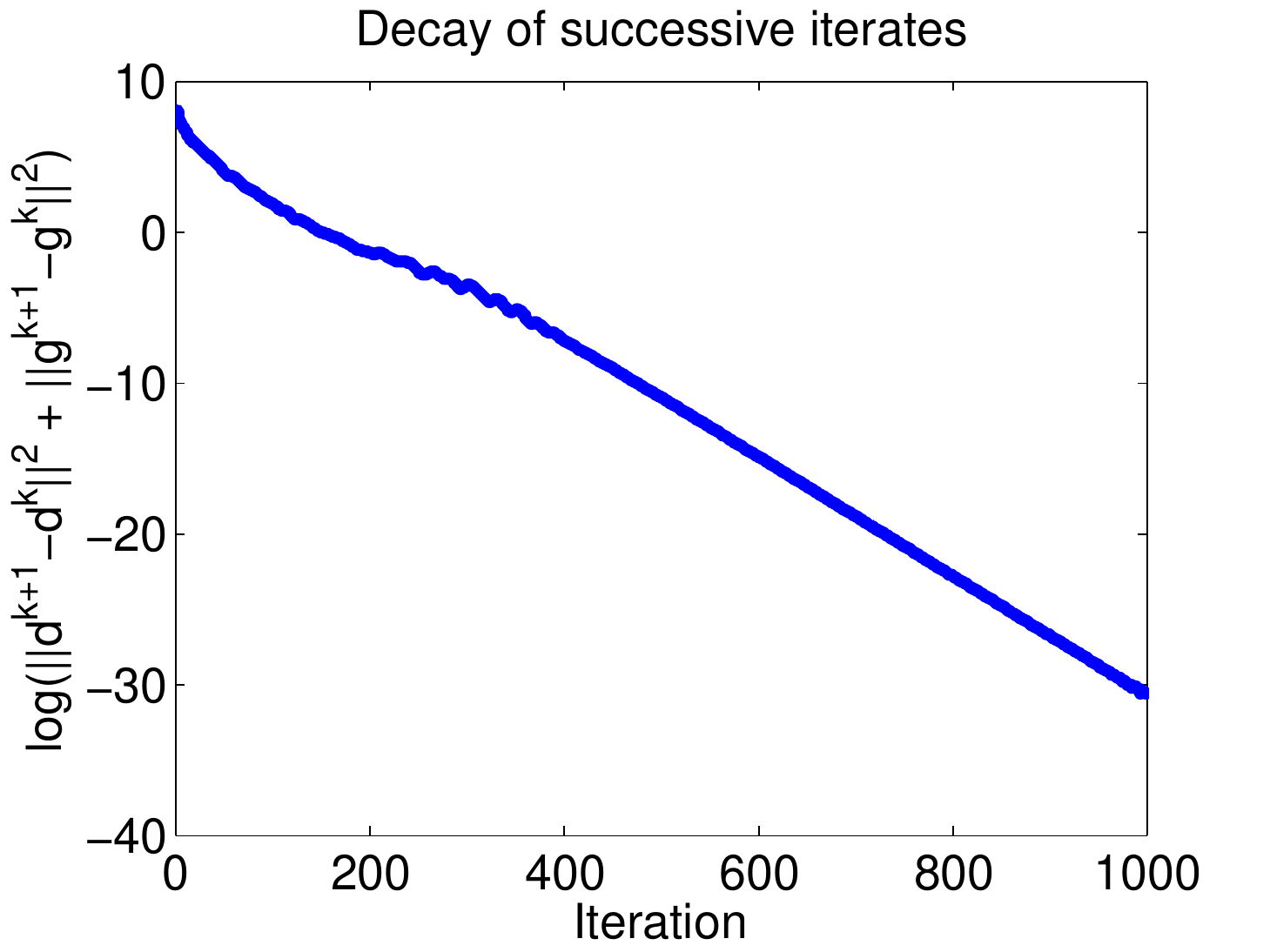}}
{\includegraphics[width=0.32\textwidth]{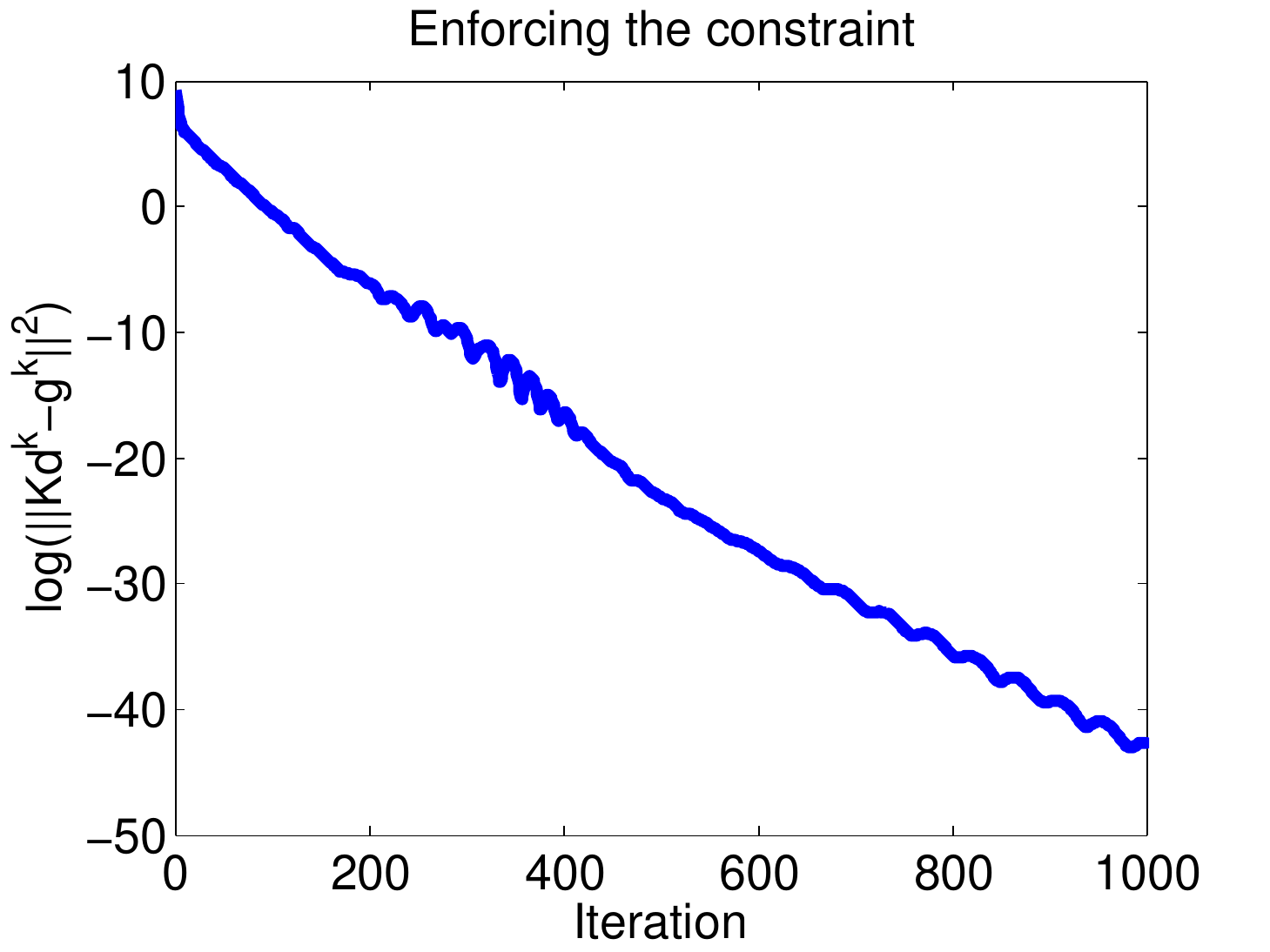}}
\caption{Convergence plots for linearized ADMM algorithm applied to the variational DFF problem. As we can see the total energy as well as the difference of successive iterates show a nice and monotonic decay. The squared $\ell^2$ norm of the difference between successive iterates decays to an order of $10^{-30}$ after the 1000 iterations. The decay of $\|Kd^k - g^k\|$ shows some small oscillations which we, however, only see for $\|Kd^k - g^k\|^2 \leq 10^{-9}$. It is remarkable that all three plots are almost linear (in the logarithmic domain) indicating a linear convergence of the displayed quantities.}
\label{fig:energyDecay}
\end{center}
\end{figure}

Starting with a small $\lambda$ gives the two variables $d$ and $g$ the freedom to behave more independent of one another. Experimentally, this helps avoiding bad local minima. Increasing $\lambda$ seems to lead to a stabilization as well as to an acceleration of the algorithm. Note that similar effects have previously been observed in the related primal-dual hybrid gradient algorithm in the nonconvex setting in \cite{Stre14}. Throughout our experiments the time step was fixed at $\tau = 8$ and we ran our algorithm for $400$ iterations. 

To give some numerical evidence of the convergence, we ran the algorithm for 1000 iterations on the ARRI data with a regularization parameter of $\alpha = 1/12$. We plotted the decay of energy in a logarithmic fashion, i.e. $\text{log}(E(d^k) - E(d^{1000}))$, in Figure \ref{fig:energyDecay} on the left hand side. As we can see the algorithm shows nice decay properties. Also included in Figure \ref{fig:energyDecay} are the decays of $\|g^{k+1}-g^k\|^2 +\|d^{k+1}-d^k\|^2$ (middle) and $\|Kd^k - g^k\|^2$ (right), both plotted in a logarithmic fashion.

 Although $\|Kd^k - g^k\|^2$ shows some oscillations in the logarithmic plot, it decayed nicely to less than $10^{-40}$. One can verify from the optimality conditions arising from the updates in \eqref{eq:linearizedADMM} that both residual converging to zero provably leads to the convergence to a critical point of our nonconvex energy (along subsequences).   
 
The total computational for determining the sharpness, determining the polynomial fitting, computing an initial depth estimate and performing $400$ iterations of the minimization algorithm took less than three seconds on a sequence of 33 images with $640 \times 480$ pixels resolution (corresponding to the image used in figure \ref{fig:variationalApproach}) using a Cuda-GPU implemetation. The source code is freely available at \url{https://github.com/adrelino/variational-depth-from-focus}. The authors would like to thank Adrian Haarbach, Dennis Mack, and Markus Schlaffer, who ported our Matlab code to the GPU. We'd like to point out that the main computational costs of the algorithm currently consists of determining the sharpness values (about 2.38 seconds in the aforementioned example) - a task which has to be done even for the classical DFF approaches which do not use regularization. The actual minimization using the ADMM algorithm takes less than 0.2 seconds. 

\section{Conclusions \& Outlook}
In this paper we proposed a variational approach to the shape from focus problem, which can be seen as a generalization of current common shape from focus approaches. It uses an efficient nonconvex minimization scheme to determine depth maps which are based on prior knowledge such as a realistic depth map often being piecewise constant. We showed in several numerical experiments that the proposed approach often yields results superior to classical depth from focus techniques. In future work we will incorporate more sophisticated regularizations in our studies. Additionally, we'll work on correcting the inherent loss of contrast caused by the total variation regularization by applying nonlinear Bregman iterations in the fashion of \cite{Bachmayr09} to our nonconvex optimization problem. %
\section*{Acknowledgements}
M.B. and C.S. were supported by EPSRC GrantEP/F047991/1, and the Microsoft Research Connections. C.S. additionally acknowledges the support of the KAUST Award No. KUK-I1-007-43. M.M. and D.C. were supported by the ERC Starting Grant "ConvexVision".

\bibliographystyle{IEEEtran}
\bibliography{references}

\begin{thebibliography}{10}
\providecommand{\url}[1]{#1}
\csname url@samestyle\endcsname
\providecommand{\newblock}{\relax}
\providecommand{\bibinfo}[2]{#2}
\providecommand{\BIBentrySTDinterwordspacing}{\spaceskip=0pt\relax}
\providecommand{\BIBentryALTinterwordstretchfactor}{4}
\providecommand{\BIBentryALTinterwordspacing}{\spaceskip=\fontdimen2\font plus
\BIBentryALTinterwordstretchfactor\fontdimen3\font minus
  \fontdimen4\font\relax}
\providecommand{\BIBforeignlanguage}[2]{{%
\expandafter\ifx\csname l@#1\endcsname\relax
\typeout{** WARNING: IEEEtran.bst: No hyphenation pattern has been}%
\typeout{** loaded for the language `#1'. Using the pattern for}%
\typeout{** the default language instead.}%
\else
\language=\csname l@#1\endcsname
\fi
#2}}
\providecommand{\BIBdecl}{\relax}
\BIBdecl

\bibitem{paolo}
P.~Favaro and S.~Soatto, \emph{3-D Shape Estimation and Image Restoration:
  Exploiting Defocus and Motion-blur}.\hskip 1em plus 0.5em minus 0.4em\relax
  Secaucus, NJ, USA: Springer New York, Inc., 2006.

\bibitem{Favaro08}
P.~Favaro, S.~Soatto, M.~Burger, and S.~Osher, ``Shape from defocus via
  diffusion,'' \emph{IEEE Transactions on Pattern Analysis and Machine
  Intelligence}, vol.~30, no.~3, pp. 518--531, 2008.

\bibitem{Lin13}
X.~Lin, J.~Suo, X.~Cao, and Q.~Dai, ``Iterative feedback estimation of depth
  and radiance from defocused images,'' in \emph{Computer Vision – ACCV
  2012}, ser. Lecture Notes in Computer Science.\hskip 1em plus 0.5em minus
  0.4em\relax Springer Berlin Heidelberg, 2013, vol. 7727, pp. 95--109.

\bibitem{Pertuz13Comparison}
S.~Pertuz, D.~Puig, M.~Garcia, and M.~Angel, ``Analysis of focus measure
  operators for shape-from-focus,'' \emph{Pattern Recogn.}, vol.~46, no.~5, pp.
  1415--1432, 2013.

\bibitem{Thelen09}
A.~Thelen, S.~Frey, S.~Hirsch, and P.~Hering, ``Improvements in
  shape-from-focus for holographic reconstructions with regard to focus
  operators, neighborhood-size, and height value interpolation,'' \emph{IEEE
  Trans. on I. Proc.}, vol.~18, no.~1, pp. 151--157, 2009.

\bibitem{Mahmood12}
M.~Mahmood and T.-S. Choi, ``Nonlinear approach for enhancement of image focus
  volume in shape from focus,'' \emph{IEEE Trans. on Image Proc.}, vol.~21,
  no.~5, pp. 2866--2873, 2012.

\bibitem{Muhammad09}
M.~Muhammad, H.~Mutahira, A.~Majid, and T.~Choi, ``Recovering 3d shape of weak
  textured surfaces,'' in \emph{Proceedings of the 2009 International
  Conference on Computational Science and Its Applications}, ser. ICCSA
  '09.\hskip 1em plus 0.5em minus 0.4em\relax Washington, DC, USA: IEEE
  Computer Society, 2009, pp. 191--197.

\bibitem{Muhammad11}
M.~Muhammad and T.~Choi, ``An unorthodox approach towards shape from focus,''
  in \emph{Image Processing (ICIP), 2011 18th IEEE International Conference
  on}, 2011, pp. 2965--2968.

\bibitem{Pertuz13}
S.~Pertuz, D.~Puig, and M.~Garcia, ``Reliability measure for
  shape-from-focus,'' \emph{Image Vision Comput.}, vol.~31, no.~10, pp.
  725--734, 2013.

\bibitem{Mahmood13}
M.~Mahmood, ``Shape from focus by total variation,'' in \emph{2013 IEEE 11th
  IVMSP Workshop}, 2013, pp. 1--4.

\bibitem{Favaro10}
P.~Favaro, ``Recovering thin structures via nonlocal-means regularization with
  application to depth from defocus,'' in \emph{Computer Vision and Pattern
  Recognition (CVPR), 2010 IEEE Conference on}, 2010, pp. 1133--1140.

\bibitem{Liu10}
H.~Liu, Y.~Jia, H.~Cheng, and S.~Wei, ``Depth recovery from defocus images
  using total variation,'' in \emph{Computer Modeling and Simulation, 2010.
  ICCMS '10. Second International Conference on}, vol.~2, 2010, pp. 146--150.

\bibitem{Namboodiri08}
V.~Namboodiri, S.~Chaudhuri, and S.~Hadap, ``Regularized depth from defocus,''
  in \emph{Image Processing, 2008. ICIP 2008. 15th IEEE International
  Conference on}, 2008, pp. 1520--1523.

\bibitem{Gaganov09}
V.~Gaganov and A.~Ignatenko, ``Robust shape from focus via markov random
  fields,'' in \emph{GraphiCon'2009}, 2009, pp. 74--80.

\bibitem{ROF}
L.~Rudin, S.~Osher, and E.~Fatemi, ``Nonlinear total variation based noise
  removal algorithms,'' \emph{Physica D}, vol.~60, pp. 259--268, 1992.

\bibitem{Nayar94}
S.~Nayar and Y.~Nakagawa, ``Shape from focus,'' \emph{IEEE Trans. Pattern Anal.
  Mach. Intell.}, vol.~16, no.~8, pp. 824--831, 1994.

\bibitem{Subbarao95}
M.~Subbarao and T.~Choi, ``Accurate recovery of three-dimensional shape from
  image focus,'' \emph{IEEE Trans. Pattern Anal. Mach. Intell.}, vol.~17,
  no.~3, pp. 266--274, 1995.

\bibitem{attouch13}
H.~Attouch, J.~Bolte, and B.~Svaiter, ``Convergence of descent methods for
  semi-algebraic and tame problems: proximal algorithms, forward-backward
  splitting, and regularized {Gauss-Seidel} methods,'' \emph{Math. Prog.}, vol.
  137, no. 1-2, pp. 91--129, 2013.

\bibitem{phamdinh}
T.~P. Dinh, H.~Le, H.~L. Thi, and F.~Lauer, ``{A Difference of Convex Functions
  Algorithm for Switched Linear Regression},'' to appear in IEEE Transactions
  on Automatic Control.

\bibitem{boyd11}
S.~Boyd, N.~Parikh, E.~Chu, B.~Peleato, and J.~Eckstein, ``Distributed
  optimization and statistical learning via the alternating direction method of
  multipliers,'' \emph{Found. Trends Mach. Learn.}, vol.~3, no.~1, pp. 1--122,
  2011.

\bibitem{zhang}
\BIBentryALTinterwordspacing
X.~Zhang, M.~Burger, X.~Bresson, and S.~Osher, ``Bregmanized nonlocal
  regularization for deconvolution and sparse reconstruction,'' \emph{SIAM J.
  Imaging Sci.}, vol.~3, no.~3, pp. 253--276, 2010. [Online]. Available:
  \url{http://dx.doi.org/10.1137/090746379}
\BIBentrySTDinterwordspacing

\bibitem{SplitBregmanConvergence}
J.~Cai, S.~Osher, and Z.~Shen, ``Split bregman methods and frame based image
  restoration,'' \emph{Multiscale Modeling $\&$ Simulation}, vol.~8, no.~2, pp.
  337--369, 2010.

\bibitem{Valkonen14}
T.~Valkonen, ``A primal-dual hybrid gradient method for non-linear operators
  with applications to mri,'' \emph{Inverse Problems}, vol.~30, no.~5, p.
  055012, 2014.

\bibitem{M14}
T.~Moellenhoff, E.~Strekalovskiy, M.~Moeller, and D.~Cremers, ``The primal-dual
  hybrid gradient method for semiconvex splittings,'' 2014, preprint. Arxiv:
  http://arxiv.org/pdf/1407.1723.pdf.

\bibitem{Li14}
G.~Li and T.~Pong, ``Splitting method for nonconvex composite optimization,''
  2014, preprint. Arxiv: http://arxiv.org/pdf/1407.0753.pdf.

\bibitem{nikolova10}
M.~Nikolova, M.~K. Ng, and C.-P. Tam, ``{Fast Nonconvex Nonsmooth Minimization
  Methods for Image Restoration and Reconstruction},'' \emph{IEEE Transactions
  on Image Processing}, vol.~19, no.~12, pp. 3073--3088, 2010.

\bibitem{HyperLaplacian}
{D. Krishnan and R. Fergus}, ``{Fast Image Deconvolution using Hyper-Laplacian
  Priors}.''\hskip 1em plus 0.5em minus 0.4em\relax {}, {2009}, pp.
  {1033--1041}.

\bibitem{attouch10}
H.~Attouch, J.~Bolte, P.~Redont, and A.~Soubeyran, ``Proximal alternating
  minimization and projection methods for nonconvex problems: An approach based
  on the kurdyka-lojasiewicz inequality,'' \emph{Math. Oper. Res.}, vol.~35,
  no.~2, pp. 438--457, May 2010.

\bibitem{Andriani13}
S.~Andriani, H.~Brendel, T.~Seybold, and J.~Goldstone, ``Beyond the kodak image
  set: A new reference set of color image sequences,'' in \emph{2013 20th IEEE
  International Conference on Image Processing (ICIP)}, 2013, pp. 2289--2293.

\bibitem{Stre14}
E.~Strekalovskiy and D.~Cremers, ``{Real-Time Minimization of the Piecewise
  Smooth Mumford-Shah Functional},'' in \emph{Proceedings of the European
  Conference on Computer Vision ({ECCV})}, 2014, p. (To appear).

\bibitem{Bachmayr09}
M.~Bachmayr and M.~Burger, ``Iterative total variation schemes for nonlinear
  inverse problems,'' \emph{Inverse Problems}, vol.~25, 2009.

\end{thebibliography}

\newpage
\section*{Appendix: Proof of Theorem \eqref{thm:convexConvergence}}
\begin{proof}
Let $\hat{d}$ be a critical point and denote $\hat{g} = K\hat{d}, \hat{b} \in \frac{\alpha \tau}{\lambda} \partial \|\hat{g}\|_{2,1}$. Furthermore, denote $d_e^k = d^k - \hat{d}$, $g_e^k=g^k-\hat{g}$, $b_e^k = b^k-\hat{b}$, $p^k_e = \nabla D(d^k) - \nabla D(\hat{d})$ and $q_e^k \in \partial \|g^k\|_{2,1} - \partial \|\hat{g}\|_{2,1}$. Then the optimality conditions arising from algorithm \eqref{eq:linearizedADMM} yield
\begin{align*}
0 =& \lambda K^T(Kd_e^{k+1} - g_e^k + b_e^k) + d_e^{k+1} - d_e^k + \tau p^k_e, \\
0 =& -\lambda (Kd_e^{k+1} - g_e^{k+1} + b_e^k) + \alpha \tau q^{k+1}_e.
\end{align*}
The inner product of the first equation with $d_e^{k+1}$ yields
\begin{align*}
0 =& \lambda \langle Kd_e^{k+1} - g_e^k + b_e^k, K d_e^{k+1} \rangle + \|d_e^{k+1}\|^2- \langle d_e^k, d_e^{k+1} \rangle  + \tau \langle p^k_e, d_e^{k+1} \rangle \\
=&  \frac{\lambda}{2}\left(\|Kd_e^{k+1} - g_e^k\|^2 + \|Kd_e^{k+1}\|^2 - \|g_e^k\|^2  \right) + \lambda \langle b_e^k, K d_e^{k+1} \rangle  + \tau \langle p^k_e, d_e^{k+1} \rangle \\
& + \frac{1}{2}\left(\|d_e^{k+1}\|^2  + \| d_e^k -  d_e^{k+1} \|^2 - \|d_e^k\|^2\right).
\end{align*}
The inner product of the second equation with $g_e^{k+1}$ results in 
\begin{align*}
0 =& -\lambda \langle Kd_e^{k+1} - g_e^{k+1} + b_e^k, g_e^{k+1} \rangle + \alpha \tau \langle q^{k+1}_e, g_e^{k+1} \rangle \\
=& \frac{\lambda}{2} \left(\|Kd_e^{k+1} - g_e^{k+1} \|^2 + \|g_e^{k+1} \|^2 - \| Kd_e^{k+1}\|^2 \right) -\lambda \langle b_e^k, g_e^{k+1} \rangle + \alpha \tau \langle q^{k+1}_e, g_e^{k+1} \rangle .
\end{align*}
Adding the two estimates above leads to 
\begin{align}
\label{eq:estimate}
0=&\frac{\lambda}{2} \big(\|Kd_e^{k+1} - g_e^{k+1} \|^2 + \|g_e^{k+1} \|^2- \|g_e^k\|^2 + \|Kd_e^{k+1} - g_e^k\|^2 \big) + \lambda \langle b_e^k, K d_e^{k+1} - g_e^{k+1} \rangle \nonumber \\
&+ \frac{1}{2}\left(\|d_e^{k+1}\|^2  + \| d_e^k -  d_e^{k+1} \|^2 - \|d_e^k\|^2\right)  + \tau \langle p^k_e, d_e^{k+1} \rangle +  \alpha \tau S_R(g^{k+1}, \hat{g}),
\end{align}
where we used the fact that $\langle q_e^{k+1}, g_e^{k+1}  \rangle =S_R(g^{k+1}, \hat{g})$. We use the update formula for $b^{k+1}$ to obtain that 
\begin{align}
\label{eq:helper1}
& \langle b_e^k, Kd_e^{k+1} - g_e^{k+1} \rangle = \frac{1}{2}(\|b_e^{k+1}\|^2 - \|b_e^{k}\|^2 - \|Kd_e^{k+1}- g_e^{k+1}\|^2).
\end{align}
Due to the linearization we have the term $\langle p_e^k, d_e^{k+1}\rangle$ in our current estimate instead of the symmetric Bregman distance. 
Thus, we estimate
\begin{align}
\label{eq:helper2}
 \langle p_e^k , d_e^{k+1} \rangle =&  S_D(d^k, \hat{d}) +  \langle p_e^k , d_e^{k+1}-d_e^k \rangle \nonumber \\
 =&  S_D(d^k, \hat{d}) - S_D(d^k, d^{k+1}) + \langle p^{k+1} , d^{k+1}-d^k \rangle -  \langle \hat{p} , d^{k+1}-d^k \rangle  \nonumber \\
 \geq&  S_D(d^k, \hat{d}) - S_D(d^k, d^{k+1})  + D(d^{k+1}) - D(d^k) -  \langle \hat{p} , d^{k+1}-d^k \rangle,
\end{align}
where we used $p^{k+1} \in \partial D(d_{k+1})$ along with the convexity of $D$ for the last inequality. 
Inserting \eqref{eq:helper1} and \eqref{eq:helper2} into \eqref{eq:estimate} yields
\begin{align*}
0\geq&\frac{\lambda}{2} \big(\|Kd_e^{k+1} - g_e^{k+1} \|^2 + \|g_e^{k+1} \|^2  - \|g_e^k\|^2 + \|Kd_e^{k+1} - g_e^k\|^2 \big)\nonumber \\
 &+ \frac{\lambda}{2}(\|b_e^{k+1}\|^2 - \|b_e^{k}\|^2 - \|Kd_e^{k+1}- g_e^{k+1}\|^2)+ \frac{1}{2}\left(\|d_e^{k+1}\|^2  - \|d_e^k\|^2\right)+ \frac{1}{2}\| d_e^k -  d_e^{k+1} \|^2\\
&- \tau S_D(d^k, d^{k+1}) + \tau  S_D(d^k, \hat{d})  +  \alpha \tau S_R(g^{k+1}, \hat{g})+ \tau (D(d^{k+1}) - D(d^k) -  \langle \hat{p} , d^{k+1}-d^k \rangle) \\
\geq&\frac{\lambda}{2} \big( \|g_e^{k+1} \|^2 - \|g_e^k\|^2 + \|b_e^{k+1}\|^2- \|b_e^{k}\|^2 \big) + \frac{\lambda}{2}\|Kd_e^{k+1} - g_e^k\|^2  + \frac{1}{2}\left(\|d_e^{k+1}\|^2  - \|d_e^k\|^2\right) \nonumber \\
&+ \tau  S_D(d^k, \hat{d})  +  \alpha \tau S_R(g^{k+1}, \hat{g})  +\tau \left(D(d^{k+1}) - D(d^k) -  \langle \hat{p} , d^{k+1}-d^k \rangle\right), 
\end{align*}
where we used the assumption $\frac{1}{2}\|d_e^{k+1} - d_e^k\|^2 - \tau S_D(p^k, p^{k+1}) \geq 0$ for the second inequality. Now we can sum over this inequality from $k=0$ to $n$ to obtain
\begin{align*}
0 \geq & \frac{\lambda}{2}(\|g_e^{n+1}\|^2 -\|g_e^0\|^2+\|b_e^{n+1}\|^2 - \|b_e^{0}\|^2)+\frac{1}{2}(\|d_e^{n+1}\|^2 - \|d_e^{0}\|^2) + \frac{\lambda}{2}\sum_{k=0}^n\|Kd_e^{k+1} - g^{k}\|^2\\
& + \tau \sum_{k=0}^n \left( S_D(d^{k}, \hat{d})+\alpha S_R(g^{k+1},\hat{g})\right) + \tau \underbrace{\left(D(d^{n+1}) - D(\hat{d}) - \langle \hat{p} , d^{n+1}-\hat{d}\rangle\right)}_{\geq 0} \\
 &- \tau \left(D(d^0)-D(\hat{d}) - \langle \hat{p},  d^0-\hat{d} \rangle\right), 
\end{align*} 
such that we finally obtain
\begin{align*}
&\frac{\lambda}{2}(\|g_e^0\|^2 + \|b_e^{0}\|^2) +  \frac{1}{2} \|d_e^{0}\|^2 +  \tau \left(D(d^0)-D(\hat{d}) - \langle \hat{p},  d^0-\hat{d} \rangle\right)\\
 \geq& \frac{\lambda}{2}\sum_{k=0}^n\|Kd_e^{k+1} - g^{k}\|^2  +  \sum_{k=0}^n \left( S_D(d^{k}, \hat{d})+\alpha S_R(g^{k+1},\hat{g})\right).
\end{align*} 
Since the sums are bounded for all $n$ and the summands are nonnegative, we can conclude their convergence to zero with a rate as least as fast as $1/k$, which yields the assumption.
\end{proof}
\end{document}